\PassOptionsToPackage{table}{xcolor}
\documentclass[sigconf,nonacm]{acmart}

\AtBeginDocument{%
  }

\setcopyright{acmlicensed}
\copyrightyear{2025}
\acmYear{2025}
\acmDOI{XXXXXXX.XXXXXXX}

\acmConference[GECCO '25]{In Genetic and Evolutionary Computation Conference }{July 14--18, 2025}{Málaga, Spain}
\acmISBN{978-1-4503-XXXX-X/18/06}

\begin{document}
\definecolor{steelblue}{rgb}{0.27, 0.51, 0.71}
\definecolor{orange}{rgb}{1.0, 0.6, 0.0}       
\definecolor{green}{rgb}{0.0, 0.5, 0.0}   


\title[Transformer Semantic Genetic Programming for Symbolic Regression]{Transformer Semantic Genetic Programming\\for Symbolic Regression}

\author{Philipp Anthes}
\email{anthes@uni-mainz.de}
\orcid{0009-0009-8383-6551}
\affiliation{%
  \institution{Johannes Gutenberg University}
  \city{Mainz}
  \country{Germany}
}
\author{Dominik Sobania}
\email{dsobania@uni-mainz.de}
\orcid{0000-0001-8873-7143}
\affiliation{%
  \institution{Johannes Gutenberg University}
  \city{Mainz}
  \country{Germany}
}
\author{Franz Rothlauf}
\email{rothlauf@uni-mainz.de}
\orcid{0000-0003-3376-427X}
\affiliation{%
  \institution{Johannes Gutenberg University}
  \city{Mainz}
  \country{Germany}
}
\renewcommand{\shortauthors}{Anthes et al.}

\begin{abstract}
In standard genetic programming (stdGP), solutions are varied by modifying their syntax, with uncertain effects on their semantics. Geometric-semantic genetic programming (GSGP), a popular variant of GP, effectively searches the semantic solution space using variation operations based on linear combinations, although it results in significantly larger solutions. 
This paper presents Transformer Semantic Genetic Programming (TSGP), a novel and flexible semantic approach that uses a generative transformer model as search operator. 
The transformer is trained on synthetic test problems and learns semantic similarities between solutions. Once the model is trained, it can be used to create offspring solutions with high semantic similarity also for unseen and unknown problems. Experiments on several symbolic regression problems show that TSGP generates solutions with comparable or even significantly better prediction quality than stdGP, SLIM\_GSGP, DSR, and DAE-GP. Like SLIM\_GSGP, TSGP is able to create new solutions that are semantically similar without creating solutions of large size. An analysis of the search dynamic reveals that the solutions generated by TSGP are semantically more similar than the solutions generated by the benchmark approaches allowing a better exploration of the semantic solution space.
\end{abstract}

\ccsdesc[500]{Software and its engineering~Genetic programming}
\ccsdesc[300]{Computing methodologies~Supervised learning by regression}
\ccsdesc[300]{Theory of computation~Program semantics}

\keywords{Genetic Programming, Transformer Models,
Semantic Operators,
Symbolic Regression}


\maketitle
\newpage
\section{Introduction}
\label{sec:Introduction}
Standard Genetic Programming (stdGP) generates solutions by applying syntactic variation operators to existing solutions in an evolutionary process. The variation operators modify the structures of previously identified solutions to create new solutions that potentially solve the target problem more effectively. However, for most problems, it is not the structure of the solutions that determines their effectiveness in solving the target problem, but rather the behavior of the solution induced by its structure. This includes symbolic regression (SR), which searches for mathematical functions that have a particular solution behavior that best captures the underlying data. This solution behavior defines which outputs a solution generates for given inputs \citep{SemanticBuildingBlocks,moraglio_geometric_2012,vanneschi_new_2013}. Unfortunately, variation operators of stdGP completely ignore how these modifications affect the behavioral properties.

To overcome this limitation, GP approaches have been developed that aim to generate new offspring that have similar solution behavior (semantics) to their predecessors \citep{Krawiec_2013_Approximating_geometric_crossover, Uy_2011_Semantically_based_crossover,moraglio_geometric_2012,vanneschi_new_2013}. These semantic-based GP methods include geometric-semantic-genetic programming (GSGP), which modifies solutions directly on their semantics through geometric-semantic operators (GSOs). Although GSOs effectively guide the search in the semantic space of solutions, they are limited by the fact that GSOs are basically linear combinations of solution structures. As a result, GSGP always creates new offspring larger than their predecessors and tends to produce rapidly (in the worst case exponentially) growing solutions of high complexity during search \citep{moraglio_geometric_2012,vanneschi_new_2013}. To mitigate this growth, variants of GSGP introduce simplification techniques, such as SLIM\_GSGP \citep{vanneschi_slim_gsgp_2024}, which continuously deflates solutions. Nevertheless, GSGP remains strongly tied to the preceding solution structures through the linear combination-based GSOs and covers only a few predefined modifications to generate similar semantics. GSGP (like most other GP variants) does not explicitly address the fact that semantics can be represented by a wide range of structurally different representations (genotypes).
Consequently, developing search operators that are able to guide the evolutionary search in the semantic space of solutions and simultaneously generate these semantic similarities through flexible structural modifications is a viable research goal.

In recent years, generative transformers have proven their ability in various domains to capture contextual relationships between sequences and generate new ones that match the desired output, regardless of their structural composition \citep{vaswani2017attention,brown2020language,briesch2024largelanguagemodelssuffer}. This paper proposes TSGP (Transformer Semantic Genetic Programming), a novel approach that considers semantic similarity within the variation process by using a generative transformer as variation operator. To apply TSGP, we train a transformer model once to identify semantic similarities between mathematical functions. After training the model, it can be used as a variation operator to sample new solutions of semantic similarity in a GP-like search. We apply TSGP to symbolic regression problems, which is a representative and relevant problem for GP approaches. We hypothesize that (1) a transformer is well suited for learning semantic similarities between mathematical functions and that (2), when integrated as a variation operator in GP, it can effectively generate functions that are semantically similar, leading to a more efficient search than the traditional approaches.

We compare the performance of TSGP with stdGP, SLIM\_GSGP and two model-based search approaches DSR and DAE-GP across five black-box SR data sets. Our results demonstrate that TSGP evolves functions with comparable or even better prediction quality on unseen data compared to the baseline methods within a fixed budget of iterations. A subsequent analysis of TSGP, stdGP, and SLIM\_GSGP shows that TSGP generates significantly smaller solutions than SLIM\_GSGP. Furthermore, the solutions generated by TSGP exhibit a higher semantic similarity than the solutions generated by stdGP. To the best of our knowledge, TSGP is the first model-based GP approach to successfully integrate the semantic similarities of solutions in the variation process. Thus, TSGP addresses the calls for variation operators that consider the semantics of solutions during variation \citep{vanneschi2014survey,moraglio_geometric_2012,krawiec2009approximating,Krawiec_2013_Approximating_geometric_crossover}.

Sect.~\ref{sec:Related Work} provides an overview of semantic approaches that incorporate semantic similarity into the variation process, as well as model-based approaches that use neural networks as search operators or to generate solutions for SR in a single forward pass. Sect.~\ref{sec:TSGP} introduces TSGP, which generates solutions with a semantic-aware transformer. Sect.~\ref{sec:Experiments} outlines the experimental design, benchmark problems, and methods. Furthermore, it presents the results of our comparative study, including additional analyses of the similarity of the generated solutions. We discuss the findings in Sect.~\ref{sec:Conclusions and Future Work} and end with concluding remarks.

\section{Related Work}\label{sec:Related Work}
\textbf{Semantic GP.} 
In recent years, several methods have been introduced that focus on the semantics of solutions. Some semantic-based GP approaches focus on the variation process by controlling the semantic similarity of new solutions. First approaches indirectly influence the variation process by discarding offspring that significantly differ in semantics from their predecessors \citep{Krawiec_2013_Approximating_geometric_crossover,Uy_2011_Semantically_based_crossover,vanneschi2014survey}. 

More advanced approaches, such as Geometric-Semantic Genetic Programming (GSGP), incorporate semantics directly into the variation process \citep{moraglio_geometric_2012}. GSGP utilizes geometric-semantic operators (GSOs), such as semantic crossover, which generates a weighted combination of parent solutions, and semantic mutation, which extends the solution by randomly generated solution structures that are guaranteed to modify the semantics of the solution in a weighted way. GSOs ensure that new solutions maintain semantic similarity. The results show that GSOs outperform standard variation operators \citep{moraglio_geometric_2012}. 
With the efficient implementation of Vanneschi et al., GSGP has proven to be applicable in a variety of real-world applications, including pharmacokinetics, financial data analysis, and several other domains \citep{Mcdermott_GSGP_Financial_Data,vanneschi_new_2013,CASTELLI_2013_ConcreteStrength,vanneschi_geometric_2014}. However, GSGP causes the offspring solutions to grow exponentially over generations. To counteract this growth, various studies have suggested simplifying the resulting offspring \citep{Martins_Solving_exponential_growth_2018, vanneschi_slim_gsgp_2024,moraglio_geometric_2012}. This includes approaches such as GSGP-Red, which simplifies solutions by aggregating repeating structures and coefficients in the solution after applying GSOs \citep{Martins_Solving_exponential_growth_2018}.
A recently published approach, SLIM\_GSGP, introduces a new deflating mechanism that reduces the size of the offspring in the variation process by subtracting a randomly selected component of the solution that was previously added to the solution \citep{vanneschi_slim_gsgp_2024}. In addition to the deflate operator, SLIM\_GSGP uses the standard operators of GSGP, which are referred to as inflate operators (semantic crossover and semantic mutation). The size of the offspring is regulated by balancing the probability of applying the inflate and deflate operators \citep{vanneschi_slim_gsgp_2024}. \\
\begin{figure*}[t]
    \centering
    \includegraphics[width=1\linewidth]{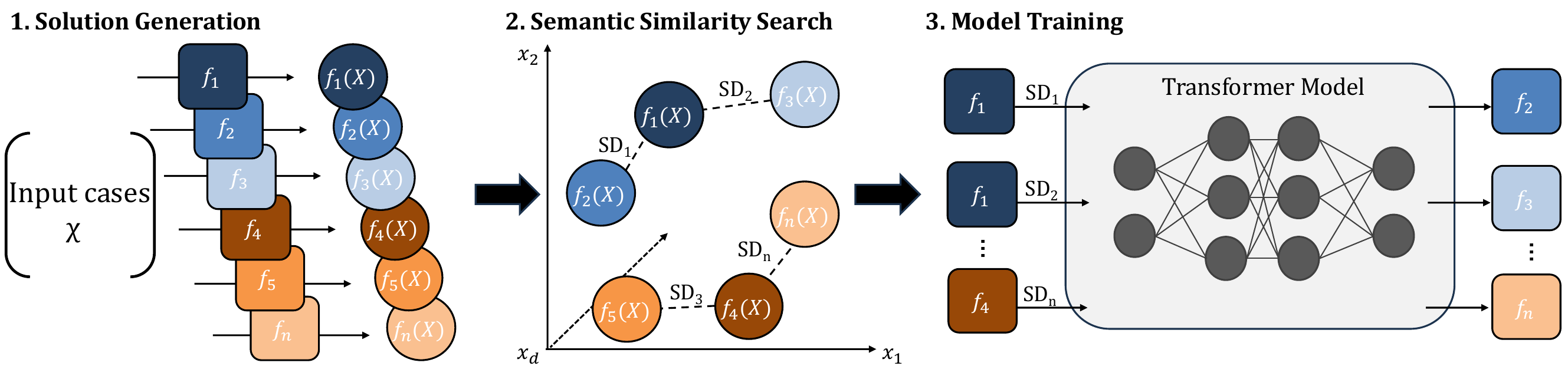}
    \caption{Model Building of TSGP. First, solutions are generated and their semantics are determined. Subsequently, solutions with similar semantics are identified in a similarity search and are passed as input-output pairs with their semantic distance ($\mathrm{SD}$) to train a transformer to generate solutions of similar semantics.}
    \label{fig:TSGP_Model}
\end{figure*}
\textbf{Neural Network-Guided Search.} 
Neural networks have been proposed as search operators by actively optimizing the trainable model parameters during the search process. Examples include Deep Symbolic Regression (DSR) \citep{petersen_deep_2021} or Denoising Autoencoder GP (DAE-GP) \citep{wittenberg_dae-gp_2020}.

DSR \citep{petersen_deep_2021} uses recurrent neural networks (RNN) to learn sampling policies through reinforcement learning (RL). The RNN auto-regressively generates solutions, which are evaluated for their fit to the underlying data. Based on a reward function, the network parameters are updated to maximize the reward, updating the sampling policy to produce solutions of higher prediction quality. In subsequent versions of DSR, several components are added to improve the quality of the solutions, such as a large-scale pre-training phase and a neural-guided GP search at the decoding stage \citep{Landajuela_2022}.

In contrast, DAE-GP utilizes a denoising autoencoder LSTM to model the distribution of solutions, which is integrated as a variation operator in GP \citep{wittenberg_dae-gp_2020}. After each selection step, the LSTM is trained to reconstruct the promising solutions by first encoding them into a latent representation and then decoding them back into their original form. Once the LSTM has learned the population's distribution, the LSTM acts as a variation operator, by applying small perturbations to solutions and passing them through the network, generating new solutions with similar properties. For SR problems, DAE-GP has demonstrated its ability to produce solutions of similar prediction quality compared to standard operators, while significantly reducing the complexity and size of the resulting structures \citep{wittenberg_small_2023}. 

\textbf{Pre-trained Neural Networks.} Besides search-based approaches, transformer-based approaches have been developed that generate solutions for SR in a single forward pass. The transformers are trained using large synthetically generated data sets consisting of numerical input values, mathematical functions, and their resulting outputs. During training, the input-output pairs are passed to the transformer, which aims to reconstruct the corresponding mathematical function token-by-token. When applied to SR problems, these approaches generate solutions by passing the input-output pairs of the analyzed data set directly to the transformer. 
Initial approaches in this area generate function skeletons with placeholders for constants, which are subsequently optimized using non-linear optimization techniques \citep{biggio_neural_2021,valipour_symbolicgpt_2021}. Subsequent approaches integrate the generation of constants directly into the decoding process, refining the sampled constants afterwards \citep{vastl_symformer_2024, kamienny2022end}. Some approaches further improve equation generation by incorporating search strategies such as Monte Carlo Tree Search (MCTS) \citep{kamienny_deep_2023,shojaee2023transformer}. For example, Transformer-Based Planning for Symbolic Regression (TSPR) \citep{shojaee2023transformer} combines a pre-trained end-to-end SR model \citep{kamienny2022end} and MCTS predicting a solution, thus actively guiding the search for functions in a more effective manner.

\section{The TSGP Approach}
\label{sec:TSGP}
Transformer Semantic Genetic Programming (TSGP) is a novel GP approach that replaces the standard variation operators of GP with a semantic-aware transformer. The trained transformer receives a parental solution as input and returns an offspring solution with similar behavioral properties. Before the transformer is incorporated into GP, it is trained in a modeling phase with a large number of synthetic pairs of similar solutions. Model building is required only once; afterwards, the trained model can be applied to unknown and unseen test problems.

\subsection{Model Building} \label{sec:Model_Building}
Model building consists of three steps, as shown in Figure \ref{fig:TSGP_Model}. The first phase involves generating a wide range of synthetic solutions and evaluating their semantics. In the second step, solutions with similar semantics are identified through a similarity search based on the semantic properties of the solutions. In phase three, solutions with similar semantics are passed to the transformer as input-output pairs for training. Each of these steps is explained in detail below.

\textbf{Solution Generation.} 
As we focus in our study on SR problems, solutions are mathematical functions \(f\), which map a set of inputs \(\mathcal{X} = \{\mathbf{x_1}, \dots, \mathbf{x_m}\}\) to a corresponding set of outputs \(f(\mathcal{X}) = \{f(\mathbf{x_1}), \dots, f(\mathbf{x_m})\}\), where each \(\mathbf{x_i} \in \mathbb{R}^d\) is a vector of \(d\) dimensions (number of features of the problem). 
To obtain a large set of mathematical functions that well cover the search space, we apply stdGP to random synthetic SR problems.\footnote{Instead of using stdGP, any other method that returns a large number of solutions covering the relevant search space is also applicable.} 
For each synthetic SR problem, stdGP returns a set of functions \(F = \{f_1, f_2, \dots, f_n\}\) where \(n\) is the total number of unique functions generated. StdGP uses double tournament selection that balances the accuracy and simplicity of generated functions, penalizing overly complex functions during the search process \citep{Sean2002FightingBloat}.
The input-output data of the synthetic SR problems are derived from randomly generated linear regression models. The data sets are standardized, so that the values of each feature have a mean of 0 and a variance of 1. To increase the diversity of the functions generated by stdGP, we apply Gaussian noise to the data sets. 

Once the set of functions $F$ is generated, we calculate the semantics of the $n$ solutions. The semantics \(s(f)\) of a function \(f\) is described by its output values $f(\mathcal{X})$, if $\mathcal{X}$ adequately covers the space of all possible inputs \citep{moraglio_geometric_2012}. 
To approximate the semantics of a solution, we evaluate each function on randomly generated standardized inputs where each of the $d$ input features has a mean of 0 and a standard deviation of 1, ensuring that the semantics comprehensively capture the function's behavior across standardized inputs.

\textbf{Semantic Similarity Search.} Two functions $f_i, f_j\in F$ are considered semantically similar if their semantics, $s(f_i)$ and $s(f_j)$, differ only slightly \citep{Uy_2011_Semantically_based_crossover}. The similarity is quantified by a semantic distance, which we define as the Euclidean distance between their semantic representations $\mathrm{SD}(f_i, f_j) = \|{ s(f_i) - s(f_j)} \|_2$ \citep{moraglio_geometric_2012}.

To identify semantically similar functions, we apply a $k$-nearest neighbor ($k$-NN) search to the set $F$. For each function \(f_i \in F\), $k$-NN returns a set of \(k\) neighbors $N_k(f_i)$ with minimal semantic distance 
\begin{equation}
N_k(f_i) = \underset{\{F' \subset F, |F'| = k\}}{\operatorname{argmin}} \sum_{f \in F'} \mathrm{SD}(s(f_i), s(f)).
\end{equation}
Thus, the semantic similarity search identifies the $k$ functions that are semantically the most similar to $f_i$. For each of the \(k\) neighbors, a pair of functions is created which is used as training data for the transformer model, where \(f_i\) serves as input and all \(f \in N_k(f_i)\) as corresponding outputs. 
The similarity search is conducted for all functions \(f_i \in F\) returning $n\times k$ possible input-output training pairs. 
The similarity search is conducted using the FAISS library \citep{douze2024faisslibrary}, which reduces computational costs by clustering semantics and measuring distances only within each cluster.

\textbf{Model Training.} The model is trained to learn what constitutes semantic similarities between functions. Thus, all input-output pairs $\{f_i,f_o\}$, which are returned by a similarity search, are used as training data for the transformer. Since mathematical functions in GP are represented as parse trees, and transformers are designed to operate on tokens, we convert each parse tree into a structured sequence of tokens to make it compatible with the input format of the transformer. This transformation is achieved by enumerating the corresponding expression tree in prefix order, traversing the tree from top to bottom and left to right \citep{wittenberg_dae-gp_2020,petersen_deep_2021}.

For the transformer model, we adopt the standard encoder-decoder architecture proposed by Vaswani et al. ~\citep{vaswani2017attention}. The encoder processes the input solution sequence $f_i$ and converts it to the latent space of the transformer. From this latent space, the decoder auto-regressively generates a new solution sequence $f_o$ token by token, guided by both the latent representation of the encoder and the previously sampled token of the sequence. During training, the model predicts the next token of the sequence based on its trainable model parameters \(\theta\). The goal of training is to identify the model parameters \(\theta^*\) that minimize the error in predicting the next token of the sequence. 
To enable the transformer to learn varying degrees of semantic similarity between solutions, we provide not only $f_i$ as input, but also the semantic distance, \(\mathrm{SD}(s(f_i),s(f_o))\) to the target output $f_o$ as an additional input to the encoder and decoder layer. 

\subsection{Model Sampling} \label{sec:Model_Sampling}
TSGP solves symbolic regression problems by using a semantic-aware transformer as a variation operator, which samples new solutions during search. Instead of applying standard variation operators of GP, TSGP uses the transformer model to create offspring solutions from parental solutions. 

Since we included the semantic distance \(\mathrm{SD}(s(f_i),s(f_o))\) as an additional input for the transformer model during training, we can adjust the semantic similarity of the generated solution by providing the desired semantic distance $\mathrm{SD^d}$ as an additional input. This allows us to control the step size and to better escape local optima. 

The transformer samples an offspring solution token-wise, based on the sequential representation of the mathematical input functions. We stop the sampling process of a solution as soon as an EOS (end-of-sequence) token is sampled or the maximum number of allowed tokens is reached. To ensure that the resulting token sequence represents a valid tree structure, we use the syntax control suggested by Wittenberg et al.~ \citep{Wittenberg:2022:EuroGP}. Syntax control modifies the probabilities of selecting tokens during the sampling process by assigning a probability of zero to any tokens that would result in an invalid tree structure.
\section{Experiments and Results}\label{sec:Experiments}
This section starts with the experimental settings including the chosen benchmark problems, the TSGP configuration and the baseline approaches used for comparison. We then evaluate the effectiveness of TSGP on five black-box regression problems and analyze its search behavior by examining the size of the solutions and the variation behavior.

\subsection{Experimental Settings}
We compare the performance of TSGP with stdGP, SLIM\_GSGP, DSR, and DAE-GP on black-box regression problems, where the ground-truth model is not known, taken from Penn Machine Learning Benchmarks (PMLB) \citep{romano2021pmlb}. PMLB is a recognized library of benchmark problems widely used in SR research, including SR bench \citep{la2021contemporary}. For our experiments, we select all available black box data sets with four features where a sufficient number of samples ($>100$) is available. 
The selected data sets include four real-world regression problems ERA (1000 samples), ESL (488 samples), Galaxy (323 samples), and LEV (1000 samples) and one synthetic data set Pollen (3848 samples). 
We standardize the data sets by scaling each feature and target variable to have a mean of 0 and a standard deviation of 1, resulting in all variables being on the same scale. Owen et al. showed that such a standardization has positive effects on prediction quality and solution size \citep{owen2018feature,dick2020feature}. 
\begin{table}
\begin{center}
\caption{Configuration parameters and values for the evaluated methods.}
\label{tab:gp_parameters}

\begin{tabular}{p{0.15\textwidth}|p{0.28\textwidth}} 
 \hline
 \rowcolor{gray!20}
 \textbf{Parameter} & \textbf{Value} \\ 
 \hline
 \rowcolor{white}
 Initialization & Ramped Half-and-Half\\
 \rowcolor{gray!10}
 Primitive Set & $\{V, ERC,+, -, \times, \%\}$ \\ 
 \rowcolor{white}
 ERC Range & [-0.5, 0.5], stepsize 0.1 \\ 
 \rowcolor{gray!10}
 Population Size & 100 \\ 
\rowcolor{white}
 Generations & 50 \\ 
\rowcolor{gray!10}
 Selection & Tournament selection of size 5 \\ 
 \rowcolor{white}
 Evaluation Metric & Root Mean Squared Error (RMSE) \\ 
\rowcolor{gray!10}
 Runs & 30 \\ 
 \hline
\end{tabular}
\end{center}
\end{table}
\begin{figure*}
    \centering
    \includegraphics[width=\textwidth]{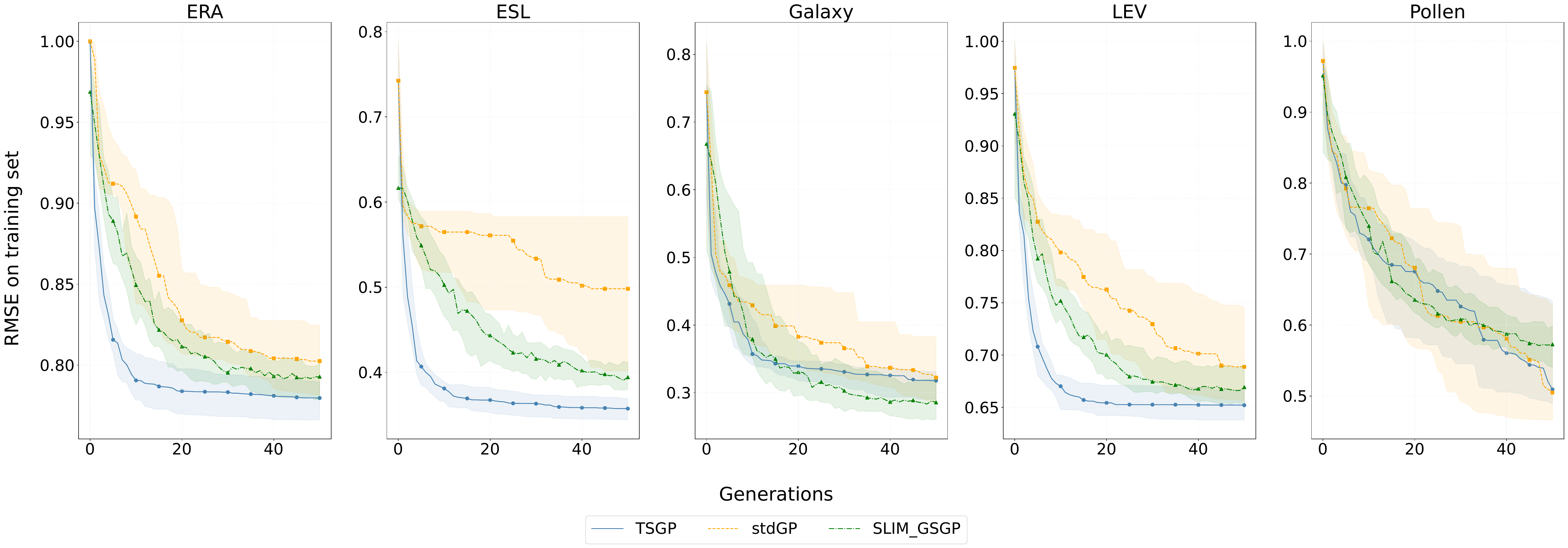} 
    \caption{Median training RMSE of the best solution of TSGP, stdGP, SLIM\_GSGP over generations for all analyzed black-box data sets.}
    \label{fig:Train-Performance}
\end{figure*}
We use the evolutionary computation framework DEAP \citep{fortin2012deap} for TSGP, stdGP, and DAE-GP. The transformer of TSGP and the LSTM of DAE-GP is implemented using Keras \citep{chollet2015keras}. For SLIM\_GSGP and DSR, we rely on the available Python implementations \citep{vanneschi_slim_gsgp_2024,petersen_deep_2021}. All general parameter settings are specified in Table \ref{tab:gp_parameters}.

For GP-based methods, initial populations are generated using Ramped Half-and-Half (RHH). For TSGP, stdGP and DAE-GP, which are based on the DEAP framework, we choose an initialization depth between 2 and 5. For SLIM\_GSGP we rely on the default initialization values as specified by the framework. We set the maximum allowed tree depth during a run to 17 ~\citep{koza1992programming}. The terminal set consists of all feasible variables $V = \{v_1, v_2, v_3, v_4\}$ that correspond to the dimensionality $d=4$ of the data set and ephemeral random constants (ERCs), which are generated within the range of [-0.5, 0.5] at a step size of 0.1. The function set consists of the binary operators addition, subtraction, multiplication, and protected division. Division by zero is avoided by replacing the resulting value with 1. The population size is set to 100 and the search is conducted for 50 generations. We use tournament selection with tournament size 5 as selection method. The data sets are partitioned into training and test sets in a 50/50 ratio. The quality of a solution is evaluated using the Root Mean Squared Error (RMSE)
\begin{equation}
\text{RMSE} = \frac{\| \mathbf{Y} - f(\mathbf{X}) \|_2}{\sqrt{m}},
\end{equation}
which is the L2 norm between the target output vector of the data set \( \mathbf{Y} \) and the output vector generated by the function \( f(\mathbf{X}) \), normalized by the total number of observations $m$ in the data set.

To create the training data for the transformer of TSGP, we apply stdGP to 50 synthetic regression problems, each consisting of four features (following the selection of the test problems). We perform stdGP runs with a population size of 2,000 until a sufficient number of functions are generated. All other GP search parameters are the same as defined in Table \ref{tab:gp_parameters}. 
As described in Sect.~\ref{sec:Model_Building}, for each generated function, its k-nearest neighbors (k=3) are determined based on their semantic distance ($\mathrm{SD}$). An input-output pair is formed for each function and one of its $k$-neighbors if their semantic similarity $\mathrm{SD} \neq0$ and $\mathrm{SD}$ < 100. With this procedure, we generate a total of 5mio semantically similar function pairs that are used to train the transformer. 

The transformer architecture is adapted from Vasawni et al.~\citep{vaswani2017attention}, with 8 attention heads and a hidden dimension of 128. We use a smaller architecture with 2 encoders and decoder stacks; the input length is limited to 100 tokens. The transformer is trained using the AdamW optimizer \citep{loshchilov2019decoupledweightdecayregularization} with a fixed learning rate of \(10^{-3}\) for 8 epochs. Systematic hyper-parameter tuning was not feasible due to the complexity of the task. It is important to note that the transformer is trained only once. The resulting trained model is then used consistently in all experiments. For model sampling, we employ $SD^d=0.1$.

StdGP uses subtree crossover and subtree mutation as variation operators. Subtree crossover is applied with a probability of 90\%, with an internal node bias of 10\% towards selecting terminal nodes \citep{koza1992programming}. Subtree mutation is applied with a probability of 10\%, where new full subtrees are generated with a depth between 0 and 2.

For SLIM\_GSGP, we use the default configuration of the framework. Thus, the SLIM+SIG2 variant is employed and no geometric crossover is performed. Variation relies solely on the semantic mutation operator. The default inflation mutation rate is set to 0.2, defining the probability of selecting the inflation mutation over the deflate operation during the mutation process of an solution.

DSR follows the default configuration of the framework. Since the results are evaluated using RMSE, the negative RMSE is chosen as a reward function. In addition, the batch size is set to match the population size of the GP-based methods (100), and the number of iterations is set to the number of generations performed (50).

DAE-GP is implemented following the specifications outlined by Wittenberg \citep{wittenberg_denoising_2023}. The denoising autoencoder LSTM features a single hidden layer that dynamically adjusts the size of the hidden dimension to match the maximum size of the solutions. Training is conducted until the training error converges, utilizing Adam optimization \citep{kingma2017adammethodstochasticoptimization} with a learning rate of $\alpha$ = 0.001. Levenshtein tree edit is used for corruption, applying a fixed edit percentage of 5\%.
\subsection{Prediction Quality}
\begin{table}
\begin{center}
\caption{Median test RMSE of the best solution identified within 50 generations for TSGP, stdGP, SLIM\_GSGP (SLIM), DSR and DAE-GP (DAE) for the five test sets. Bold values denote the best prediction quality (lowest RMSE). Significant differences of the best results are indicated by the label symbols.}
\label{tab:prediction_results}
\begin{tabular}{l|r|r|r|r|r} 
 \hline
 \rowcolor{gray!20}
\textbf{Data set} & \textbf{\(_a\text{TSGP}\)} & \textbf{\(_b\text{stdGP}\)} & \textbf{\(_c\text{SLIM}\)} & \textbf{\(_d\text{DSR}\)} & \textbf{\(_e\text{DAE}\)}\\ 
 \hline
  \rowcolor{white} {ERA} & \textbf{\(_{bde}\text{0.797}\)} & 0.817 & 0.810   & 0.852& 0.904\\ 

 \rowcolor{gray!10} {ESL} & \textbf{$_{bcde}\text{0.379}$} & 0.502 & \(\text{0.418}\) & 0.507 & 0.595\\ 

 \rowcolor{white} {Galaxy} & 0.327 & 0.337 & \textbf{$_{de}\text{0.305}$} & 0.434 & 0.468\\ 

 \rowcolor{gray!10} {LEV} & \textbf{\(_{bde}\text{0.672}\)} & 0.703 & 0.681 & 0.773 & 0.842\\ 

 \rowcolor{white} Pollen & 0.518& \textbf{$_{de}\text{0.514}$} & 0.569 & 0.750 & 0.752\\ 

  \hline
\end{tabular}
\end{center}
\end{table}

\begin{figure*}
    \centering
    \includegraphics[width=\textwidth]{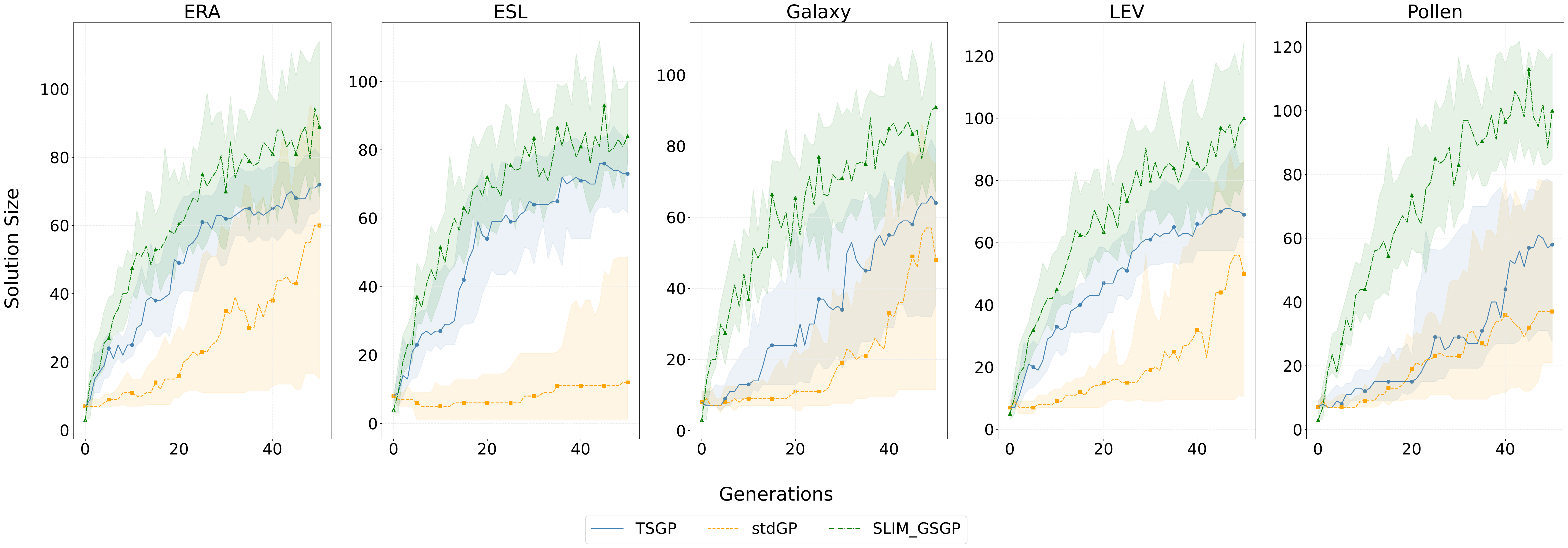} 
    \caption{
    Median solution size of the best solution of 
    TSGP, stdGP, SLIM\_GSGP over generations for all analyzed black-box data sets.
    \label{fig:Size}
    }
\end{figure*}
We analyze the prediction quality of TSGP and the baseline methods for the selected SR problems by calculating the RMSE of the best solutions, where a lower RMSE indicates a better prediction quality. The best performing solution on the training set is used to compute the RMSE on the test set. The test results for all methods across all standardized datasets are summarized in Table \ref{tab:prediction_results}. These values represent the median of 30 independent runs. Bold values indicate the highest prediction quality, which are tested for statistically significant differences using the Wilcoxon rank sum test with a significance level of $\alpha = 0.05$. The method labels ($a$,$b$,$c$,$d$,$e$) highlight statistically significant differences \cite{geiger2023down}.

We find that TSGP outperforms all baseline methods, including stdGP, SLIM\_GSGP, DSR, and DAE-GP, on the majority of the evaluated data sets. For the data sets ERA, ESL and LEV, the solutions generated by TSGP are of significantly higher prediction quality than those generated by stdGP, DSR and DAE. Especially on the ESL data set, TSGP performs significantly better than all other evaluated baselines.
On the Galaxy dataset, SLIM\_GSGP achieves solutions with the highest prediction quality, while for Pollen, stdGP generates the best solutions. However, for both datasets, no significant differences to TSGP can be observed.
The model-based approaches DSR and DAE-GP achieve significantly worse results on all data sets examined. Therefore, the remaining analysis focuses only on the comparison of TSGP with stdGP and SLIM\_GSGP. We attribute the poor performance of DSR and DAE-GP to the fact that both adapt their neural networks within the search to find good solutions and, therefore, depend on a large amount of training data during the search. In contrast, TSGP is trained only once during a model-building phase on synthetic problems and can transfer its knowledge of useful variations to unseen datasets without requiring additional training.

Our results confirm the findings of Vanneschi \citep{vanneschi_slim_gsgp_2024} that semantic-based methods outperform standard operators in terms of prediction quality. Both TSGP and SLIM\_GSGP consistently outperform stdGP across all datasets, except for Pollen, where TSGP achieves only comparable results. These results support the hypothesis that incorporating semantic operators into the variation process increases the effectiveness of the GP search and leads to improved prediction quality of the identified solutions.

To provide deeper insights into the search behavior of TSGP, we analyze the prediction quality of the best solutions over generations. Figure \ref{fig:Train-Performance} shows the median RMSE of the training set (of 30 independent runs) over the number of generations. Additionally, the interquartile range (IQR) is included to quantify its variability. Note that the initial RMSE values of the first generation slightly differ between TSGP and stdGP compared to SLIM\_GSGP due to differences in their frameworks and initialization.

The results indicate that TSGP finds high-quality solutions significantly faster than stdGP or SLIM\_GSGP on the majority of data sets. This is demonstrated by ERA, ESL, and LEV, for which TSGP finds solutions with high prediction quality after only 10 generations, whereas the comparison methods converge noticeably slower. Especially on ESL, TSGP achieves a significantly lower RMSE after 10 generations (0.381) than stdGP (0.565) or SLIM\_GSGP (0.503). On the Galaxy and Pollen data sets, TSGP converges at a rate similar to stdGP and SLIM\_GSGP, showing no differences in convergence between the methods.

The IQR of TSGP across all data sets is noticeably smaller than stdGP and is similar in magnitude to that of SLIM\_GSGP. This reduced spread indicates less variability in the prediction quality of the best solutions across the conducted runs. These results suggest that semantic-based methods, such as TSGP, are more stable in their exploration of the solution space. Unlike stdGP, their variation operators rely less on random perturbations of the solution behavior, leading to more consistent and reliable results over multiple runs.
\begin{figure*}
    \centering
    \includegraphics[width=\textwidth]{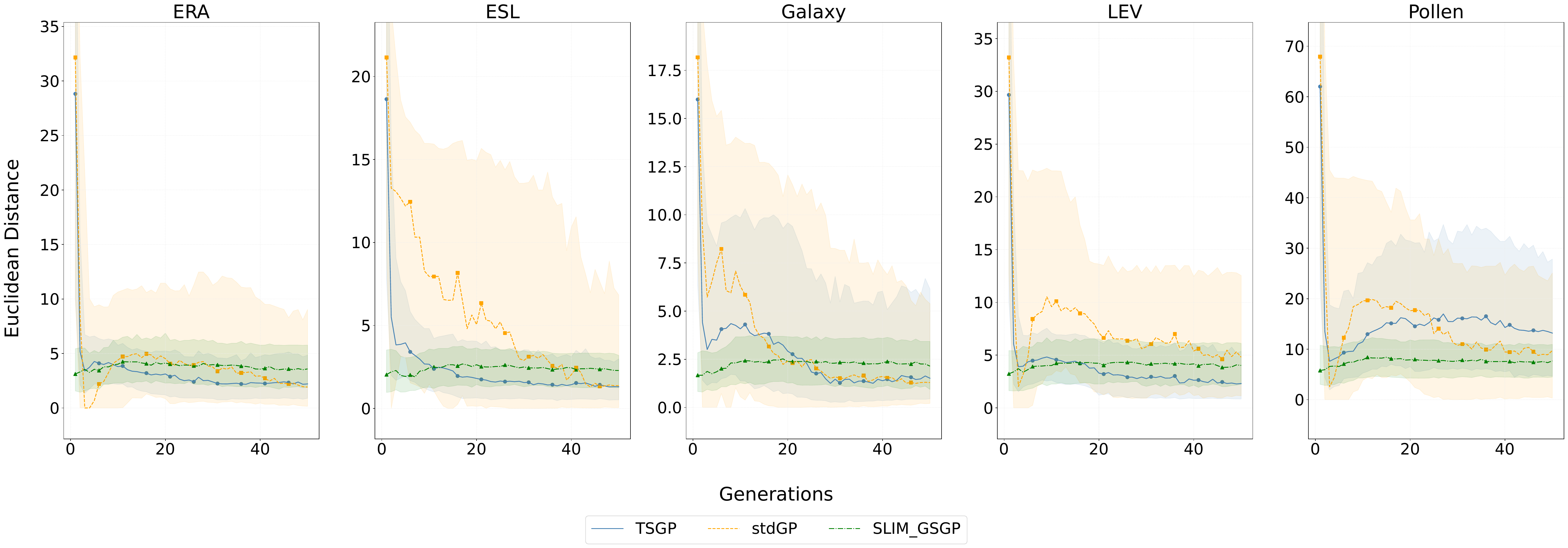} 
    \caption{Median Euclidean Distance between the semantics \(s(f_i)\) and \(s(f_o)\) for TSGP, stdGP, SLIM\_GSGP over generations for all analyzed black-box data sets.}
    \label{fig:Euclidean-Distance}
\end{figure*}
\subsection{Solution Size}
\begin{table}
\begin{center}
\caption{Median solution size of the best solution identified within 50 generations by TSGP, stdGP and SLIM\_GSGP (SLIM). Values in bold indicate the smallest solutions. Significant differences with respect to all other methods are indicated by the label symbols.}
\label{tab:solution_size} 
\begin{tabular}{l|r|r|r} 
 \hline
 \rowcolor{gray!20}
\textbf{Data set}& \textbf{\(_a\text{TSGP}\)} & \textbf{\(_b\text{stdGP}\)} & \textbf{\(_c\text{SLIM}\)} \\ 
 \hline
  \rowcolor{white} ERA & \(_c\text{72}\) & \textbf{\(_c\text{60}\)} & 89 \\ 
 \rowcolor{gray!10} ESL & \(_c\text{73}\) & \textbf{\(_{ac}\text{12}\)} & 84 \\ 
 \rowcolor{white} Galaxy & \(_c\text{64}\) & \textbf{\(_c\text{48}\)} & 91 \\ 
  \rowcolor{gray!10} LEV & \(_c\text{69}\) & \textbf{\(_c\text{50}\)} & 100 \\ 
  \rowcolor{white} Pollen & \(_c\text{58}\) & \textbf{\(_c\text{37}\)} & 100 \\ 
  \hline
\end{tabular}
\end{center}
\end{table}
Usually, not only prediction quality, but also size of the solutions is a decisive criterion. Simpler and more compact solutions are generally preferred, as large solutions tend to be more complex and can make human interpretation difficult \citep{wittenberg_small_2023,vanneschi_slim_gsgp_2024,Sean2002FightingBloat}. Consequently, we quantify the solution size by counting the number of nodes of the best solutions. Table \ref{tab:solution_size} lists the solution sizes of TSGP, stdGP, and SLIM\_GSGP. As before, these values represent the median of 30 independent runs. Statistically significant differences are examined between all the evaluated methods and are marked with the respective labels ($a$,$b$,$c$). The smallest solution sizes are highlighted in bold.

We find that the best solutions found by stdGP have the smallest size, followed by TSGP and SLIM\_GSGP. The small sizes of stdGP's solutions are expected, since their prediction quality is worse than those of the semantic methods, except on the Pollen data set. A comparison of the solution sizes between stdGP and the semantic methods shows statistically significant differences between stdGP and SLIM\_GSGP, with all solutions of SLIM\_GSGP being significantly larger than those of stdGP. However, no statistical differences in solution size were observed between TSGP and stdGP.

Interestingly, TSGP consistently finds statistically significantly smaller solutions than SLIM\_GSGP across all data sets. This is particularly noteworthy because TSGP also generates solutions with higher prediction quality than SLIM\_GSGP for most data sets. Compared to the geometric semantic operators of SLIM\_GSGP, the transformer is able to generate semantic similarities through flexible structural modifications. Being trained on semantically similar pairs of solutions of different structural compositions, the semantic similarities can be generated by the transformer through various structural modifications. Therefore, TSGP does not depend on generating increasingly larger solutions within the search, but can generate semantically similar solutions of flexible structure and length. This is reflected in significantly smaller solutions generated across all data sets compared to SLIM\_GSGP. 

This is also evident when studying the sizes of the best solutions over generations, as shown in Figure \ref{fig:Size}. During the search, the best solutions found by TSGP are always smaller than the best solutions from SLIM\_GSGP, although TSGP's solutions show higher prediction quality for the majority of the data sets. 
For the data sets where TSGP achieves higher prediction quality than the comparison methods (ERA, ESL, LEV), TSGP quickly identifies large solution structures with high prediction quality. However, for the data sets Galaxy and Pollen, where the solution quality of TSGP is comparable to stdGP and SLIM\_GSGP, the increase in solution size for TSGP is more moderate and comparable to the increase for stdGP and is significantly smaller than Slim\_GSGP with a decrease in solution size up to 42\% on Pollen.
\subsection{Variation behavior}
To better understand the variation behavior of TSGP, we analyze the solutions sampled during the search by comparing each solution \(f_i\) with its offspring \(f_o\) over generations. We focus on the similarity of the solution behavior by measuring the distance between their semantics, $s(f_i)$ and $s(f_o)$, using the Euclidean distance. The semantics are approximated from the output of the functions when applied to the test set of the target problem. We only consider successful variations where the offspring is structurally different from its parent, which is almost always the case. Figure \ref{fig:Euclidean-Distance} plots the median distances of the variations over generations. The results are for 30 independent runs, with the IQR provided to analyze the variability of the semantic distances. Note that the magnitude of the Euclidean distance depends on the magnitude of the test set. Semantic similarities can therefore only be compared within a single data set, not across data sets.

For almost all data sets (except Pollen), TSGP's semantic-aware transformer produces offspring with the highest semantic similarity compared to stdGP and SLIM\_GSGP. This is indicated by the fact that the median semantic distance for TSGP is lower than for stdGP or SLIM\_GSGP for most generations. These results highlight the ability of transformers to recognize semantic variations of solutions during training and their ability to generate new solutions that exhibit similar behavioral characteristics.

Particularly noticeable is that the IQRs of the semantic methods are significantly smaller than those of stdGP with its syntactically based variation mechanisms. The semantic methods thus exhibit a more stable variation with regard to solution behavior. In particular, the 75th percentile of stdGP is significantly higher than that of the semantic methods. This indicates that the standard operators of GP often generate offspring that differ greatly from the solution behavior of the given input solution. This is to be expected, since stdGP with mutation and crossover does not incorporate the behavior of solutions into the variation process, and thus often generates solutions that differ greatly in solution behavior. Semantic variation operators such as TSGP, on the other hand, consistently create solutions with more similar solution behavior in the majority of the population, and thus transfer the favorable properties of the solutions more steadily to their offspring. 

There are differences between the methods in the progression of behavioral similarity over generations. Especially at the beginning of the search, both TSGP and stdGP generate variations of high semantic distance. This is due to the random initialization of the population, which consists of solutions with very diverse solution behavior. These solutions initially have a low fit to the underlying problem, which means that TSGP, which is trained on solutions for synthetic SR problems, is unlikely to be able to generate solutions of low semantic distance to these random solutions at the beginning of the search.
For data sets where TSGP can identify solutions with high prediction quality in a few generations, the semantic distance is significantly lower compared to the other methods. In these cases, TSGP effectively explores the search space and generates new solutions that retain useful properties from previous solutions.
On the data sets Galaxy and Pollen, the semantic distance of TSGP is comparable to that of stdGP. Furthermore, the 75th percentile of these data sets is not as low as compared to stdGP as it is on the other data sets. This could explain why the prediction quality for these data sets is only comparable between TSGP and the other methods and does not outperform the prediction quality as on ERA, ESL and LEV. Solutions are sought for these data sets, on which the transformer can generate only semantically similar structures to a limited extent. As a result, the search is not as effective as on the other data sets.

\section{Conclusions and Future Work}\label{sec:Conclusions and Future Work}
We introduced TSGP, a novel semantic-based GP approach that uses a semantic-aware transformer as a variation operator. For TSGP, a transformer is trained once on synthetic data. The trained transformer can then be used as a search operator, which is able to generate favorable and semantic variations for unseen data sets.

For the majority of the analyzed black-box regression data sets, TSGP significantly outperforms stdGP, SLIM\_GSGP, DSR and DAE-GP in prediction quality and converges after only a few generations to high-quality solutions, significantly faster than the baseline methods. Furthermore, the solutions generated by TSGP are significantly smaller than those of SLIM\_GSGP, with reductions up to over 40\%, while maintaining solution quality that is comparable or even significantly better. We assume that this is due to the fact that the semantic-aware transformer in TSGP is more flexible in generating new semantic variations than SLIM\_GSGP, where variations are based on linear combinations of solution structures. An analysis of the variation behavior shows that the TSGP is able to generate offspring with solution behavior that is significantly more similar to the behavior of standard GP operators.

As next steps, we will focus on extending TSGP to data sets of varying dimensionality to enable it to be applied to a wider range of data sets. Furthermore, we study the effects of training data on TSGP's variation process as we believe that a greater variety of training data could have a positive impact on the generalizability of TSGP's high prediction quality across data sets. We also plan to systematically control the step size of the transformer during the search as larger step sizes $SD^d$ would allow us to better escape local optima.
\begin{acks}
We would like to thank David Wittenberg for the valuable insights into DAE-GP and for providing his framework. We also like to thank Alina Geiger, Martin Briesch and the entire team in Mainz for the inspiring discussions and thoughtful contributions.
\end{acks}
\bibliographystyle{ACM-Reference-Format}
\bibliography{TSGP}

\end{document}